\documentclass{article}

\usepackage{PRIMEarxiv}
\usepackage{float}
\usepackage[utf8]{inputenc} 
\usepackage[T1]{fontenc}    
\usepackage{hyperref}       
\usepackage{url}            
\usepackage{booktabs}       
\usepackage{amsfonts}       
\usepackage{nicefrac}       
\usepackage{microtype}      
\usepackage{lipsum}
\usepackage{fancyhdr}       
\usepackage{graphicx}       
\usepackage{enumitem}
\usepackage{algorithm}
\usepackage[table]{xcolor}
\usepackage{algpseudocode}
\usepackage{amsmath}
\newcolumntype{M}[1]{>{\centering\arraybackslash}m{#1}}
\graphicspath{{media/}}     

\title{Deep Image Restoration For Image Anti-Forensics 
}

\author{
  Eren Tahir\\
  Yildiz Technical University \\
  Istanbul, Turkey\\
  \texttt{eren.tahir@std.yildiz.edu.tr} \\
    \And
  Mert Bal \\
  Yildiz Technical University \\
  Istanbul, Turkey\\
  \texttt{mertbal@yildiz.edu.tr} \\
}

\begin{document}
\maketitle

\begin{abstract}
While image forensics is concerned with whether an image has been tampered with, image anti-forensics attempts to prevent image forensics methods from detecting tampered images. The competition between these two fields started long before the advancement of deep learning. JPEG compression, blurring and noising, which are simple methods by today's standards, have long been used for anti-forensics and have been the subject of much research in both forensics and anti-forensics. Although these traditional methods are old, they make it difficult to detect fake images and are used for data augmentation in training deep image forgery detection models. In addition to making the image difficult to detect, these methods leave traces on the image and consequently degrade the image quality. Separate image forensics methods have also been developed to detect these traces. In this study, we go one step further and improve the image quality after these methods with deep image restoration models and make it harder to detect the forged image. We evaluate the impact of these methods on image quality. We then test both our proposed methods with deep learning and methods without deep learning on the two best existing image manipulation detection models. In the obtained results, we show how existing image forgery detection models fail against the proposed methods. Code implementation will be publicly available at https://github.com/99eren99/DIRFIAF .
\end{abstract}


\section{Introduction}
Image forensics is a field that deals with determining whether an image is authentic or not, and whether its content and features have been altered. Although it is a field that has gained momentum with the breakthrough of deep learning, it is a field that has been studied since long before. In the early 2000s, both deep learning\cite{4054529} and non-deep learning\cite{1329901} statistical methods were developed. One of the pioneering models in image forgery detection after the deep learning breakthrough is Mantranet\cite{8953774}. Before Mantranet, most of the studies in this field were statistical methods. Contrary to the past, the methods developed after Mantranet are mostly deep learning methods. The reason for this change is that traditional (statistical) methods lag behind deep learning methods in terms of success. Another reason is that statistical methods are specialized methods. Examples for specialized methods: detection of traces left by JPEG compression\cite{5946978}, detection of noise differences\cite{MAHDIAN20091497}, detection of CFA mismatches\cite{6210378}. Deep learning can handle all these features at the same time with its blackbox structure.

The opposite field of image forensics, image anti-forensics, aims to prevent a forged image from being detected by image forensics methods. Like statistical image forensics methods, image anti-forensics methods are often not generic.Studies have been carried out to remove the camera trace\cite{9156605} , to remove the traces formed after JPEG compression\cite{9306490} and anti-forensics methods for median bluring and contrast enhancement\cite{SHARMA2020102682}. In a recent study, the denoising diffusion model, which is also used in adverserial attacks, was used for image anti-forensics\cite{Tailanian_2024_WACV}.

In the training of image manipulation detection models, simple anti-forensics methods such as blurring, rescaling, resizing, noising, JPEG compression, random cropping are used as data augmentation methods to increase the robustness of the models. While there is no existing dataset produced using advanced anti-forensics methods, using only simple anti-forensics methods in model training makes the models ineffective against anti-forensics methods. The aforementioned simple methods not only make it difficult to detect image manipulations, but also leave traces in the image that degrade the image quality. In a time when high-resolution cameras are accessible, a low-quality image is inherently suspect in terms of authenticity. In our study, we mitigate the quality degradation caused by simple methods with deep image restoration methods. To the best of our knowledge, no previous work has used deep image restoration models for this purpose at this scale. In a 2015 study using image restoration in the field of anti-forensics, the median filter and JPG compression were addressed with non-deep learning methods\cite{fan:tel-01216243}. In a recent image forgery detection study, model's robustness is tested with Adobe Photoshop's JPEG artifact removal neural filter \cite{karageorgiou2024fusion}.  

Image restoration deals with the removal of undesirable features from a damaged or damaged image and the restoration of a clean image.  

\subsection{Image Restoration Tasks for the Enhancement of Images Applied to Anti-Forensics Methods}
The field of image restoration has tasks that are the opposite of the mentioned simple anti-forensics methods. With these methods, the image quality degradation caused by simple anti-forensics methods can be reduced.
\begin{itemize}[leftmargin=*]
\item \textbf{Image Deblurring for blurring:} Image debluuring aims to remove blur from the image.
\item \textbf{Image Super-Resolution for rescaling:} Image super-resolution aims to increase the resolution while preserving the quality of the image.
\item \textbf{Image Denoising for noising:} Image denosing aims to remove noise from the image.
\item \textbf{JPEG Compression Artifact Removal for JPEG compression:}  JPEG compression artifact removal task aims to remove traces left by JPEG compression in the image that degrade the image quality.
\end{itemize}

\section{Related Work}
\subsection{Image Forgery Detection Models and Datasets}
Image forgery detection is the task of determining at the image level whether an image is genuine or tampered. Image forgery localization is the task of determining at the pixel level whether an image is real or falsified. Trufor\cite{guillaro2023trufor} is an image forgery detection and localization deep learning model that both utilizes RGB and noise features. It extracts noise features via a convolutional network called Noiseprint++. Noiseprint++ is trained with contrastive learning and has objective to exract noise residual of an image. Noiseprint++ ,unlike it's ancestor Noiseprint\cite{8713484}, is durable to image's editing history. An other outstanding image forgery model is Early Fusion\cite{triaridis2024exploring}. Early Fusion employs previous succesful methods at the same time. Those methods are Noiseprint++, Bayar Convolution\cite{10.1145/2909827.2930786} and SRM high pass filters\cite{6197267}. Combining those successful methods resulted in an outsanding accuracy improvement over benchmark datasets. Two of those benchmark datasets that we have used in thus study are COVERAGE\cite{Wen2016COVERAGEA} and DSO-1\cite{6522874}. COVERAGE dataset is a dataset for copy move forgery including 100 tampered and 100 authentic images. COVERAGE has small resolution images around 400x400. Early Fusion has the best forgery detection score on this dataset, where as Trufor has the third best score\cite{paperswithcodePapersWithCov}. DSO-1 dataset is a dataset for splicing forgery including 100 tampered and 100 authentic images. DSO-1, unlike COVERAGE, has high resolution images greater than 1500x1500. Early Fusion has the best forgery detection score on this dataset, where as Trufor has the second best score\cite{paperswithcodePapersWithDso}.

\subsection{Image Restoration Methods}
Image restoration is an image processing task that removes defects and restores the clean image from its degraded state to improve image's visual quality. Image restoration methods are divided into blind and non-blind. Non-blind methods are methods that know exactly the cause of the distortion in the image and improve the image accordingly. Blind methods, on the other hand, are methods that improve the image without having clear information about the distortion in the image. Image restoration task has subtasks such as image denoising, image dehazing, image deblurring, JPEG artifacts removal and image super-resolution. Some of these subtasks can be used to eliminate the image quality degradation caused by anti-forensics methods. FBCNN\cite{jiang2021towards} is a JPEG compression artifacts removal network. It is a blind method that does not need to know quality factor of JPEG compression applied to image. FBCNN can restore the JPEG compressed image and predict the quality factor of compression at the same time. Restormer\cite{Zamir2021Restormer} is a image restoration architecture that can be used on several image restoration subtasks. Model was designed considering high computational complexity of transformer architecture so that model can be used with high resolution images.Tasks that architecture is used on are image deraining, single-image motion deblurring, defocus deblurring (single-image and dual-pixel data), and image denoising (Gaussian grayscale/color denoising, and real image denoising). SwinIR\cite{liang2021swinir} is a image super-resolution deep learning model adopting the window-based and local attention strategy. This strategy uses limited receptive field in order to reduce computational complexity. SwinFIR\cite{zhang2022swinfir} study improved the SwinIR by using fast fourier convolution components which resulted in increased receptive field and more global information captured.

\subsection{Image Quality Metrics}
Since it is not realistic to analyze and interpret the image quality of thousands of converted images individually, we used image quality metrics to measure the quality of the generated images. Full-Reference Image Quality Assessment methods use the original image to measure the quality of a degraded image. PSNR is a full reference IQA metric which uses mean squared error between distorted and original image. SSIM\cite{1284395} is also a full reference IQA which takes into account dynamics of human visual system unlike PSNR. SSIM takes into account structure, contrast and luminance features while calculating similarity between images. There is an another type of IQA that is called blind (no reference) IQA which uses only the distorted image to measure image quality. BRISQUE\cite{6272356} is a no reference IQA metric. BRISQUE score is calculated via a support vector regression model which takes visual features as input. Higher PSNR and SSIM mean better image quality, while lower BRISQUE score means better image quality.

\section{Proposed Anti-Forensics Methods}
Unlike other image anti-forensics methods we use a two step approach that corruption followed by restoration. In addition to methods with deep learning, we use methods without deep learning to find out whether the computational overhead of deep learning methods is worth the effort. Experimented methods:
\begin{enumerate}
  \item \textbf{Blur\&Sharp:} It is one of our two methods without deep learning. First, we use Gaussian blurring with kernel size 5. The effect of Gaussian blur on image forgery detection models can be seen in Figure 9 from Trufor's paper. At first, we tried to remove the blur with image blurring models of image restoration. None of the methods we tried gave satisfactory results. Then the Wiener filter was tried and again the results were not good enough. Finally, the sharpening filter(with kernel size 3), which is the reverse process of blurring, was tried and satisfactory results were obtained. Look Figure \ref{fig:1} for kernels.
  \begin{figure}[H]
      \begin{minipage}{0.5\linewidth}
        \centering
        \[1/273\left[
        \begin{matrix}
          1 & 4 & 7 & 4 & 1\\
          4 & 16 & 26 & 16 & 4\\
          7 & 26 & 41 & 26 & 7\\
          4 & 16 & 26 & 16 & 4\\
          1 & 4 & 7 & 4 & 1
        \end{matrix}
        \right]\]
        5x5 Gaussian Blur kernel
      \end{minipage}
      \begin{minipage}{0.5\linewidth}
        \centering
        \[\left[
        \begin{matrix}
        0 & -1 & 0\\
        -1 & 5 & -1\\
        0 & -1 & 0
        \end{matrix}
        \right]\]
        3x3 Sharpening kernel
      \end{minipage}
      \caption{Kernels of Blur\&Sharp method}
    \label{fig:1}
    \end{figure}
    
  \item \textbf{Downsize\&Upsize:} In our previous study\cite{tahir2024deep} we observed a drastic accuracy drop on forgery detection when images are resized to 512x512 resolution. Inspired by this, we downsized  images asymmetrically at half the width and three-quarters the height. As a restoration step, we upsized to their original size. "LANCZOS4" interpolation of OpenCV library was used for downsizing and "INTER CUBIC" interpolation of OpenCV library was used for upsizing. This is our last method without deep learning.  
  \item \textbf{JPEG Compression\&JPEG Compression Artifacts Removal:} This method first compress image with JPEG compression at a fixed quality factor. After corruption, image is restored with blind JPEG compression artifact removal model FBCNN. 50 and 70 quality factors of OpenCV library were experimented for this method. Unlike robustness test in OMGFuser paper, we additionally compress images before artifact removal process\cite{karageorgiou2024fusion}. 
  \item \textbf{Gaussian Noise\&Denoise:} Gaussian noise with a fixed sigma is added to images. As a restoration step, Restormer's pretrained non-blind Gaussian denoisers are used to remove noise from images. 15 and 25 sigma values were experimented for this method. The difference with the method using Diffusion models for anti-forensics\cite{Tailanian_2024_WACV} is that the main goal of this task is to directly remove noise while preserving image content, whereas diffusion models aim to create new images. Due to memory limitations, overlapping sliding windows were used in inference for this method (see Github repository).
  \item \textbf{Downscale\&Upscale:} This method employs image super-resolution task of image restoration. Images are resized so that their width and height are reduced by half with "LANCZOS4" interpolation of OpenCV library, in other words they are 2x downscaled. Downscaled images are than upscaled to their original size via pretrained SwinFIR 2x upscaler model.
\end{enumerate}

\subsection{Image Forgery Detection Scores}
Image forgery detection is an image level classification task that predicts if an image is forged or authentic. In this binary classification objective; positive class represents forged images and negative class represents authentic images. On the forensics side, the analyst aims to correctly classify all images, i.e. to increase Accuracy value, while on the anti-forensics side, the manipulator aims to reduce the detection of fake images by the analyst, i.e. to reduce Recall value.

\begin{itemize}[leftmargin=*]
\item \textbf{True Positive:} Correctly classified forged images.
\item \textbf{True Negative:} Correctly classified authentic images.
\item \textbf{False Positive:} Misclassified authentic images.
\item \textbf{False Negative:} Misclassified forged images.
\item \textbf{Accuracy ( TP + TN ) / ( TP + TN + FP + FN ) :} Ratio of correctly classified images to all images.

\item \textbf{Recall ( TP ) / ( TP + FN ) :} Ratio of correctly classified fake images to all forged images.
\end{itemize}

\begin{table}[H]
\setlength\arrayrulewidth{1pt} 
\renewcommand{\arraystretch}{1.4}
\centering
\begin{tabular}{|r|c|c|c|c|c|c|}
\hline
\rowcolor{lightgray}\multicolumn{1}{|c|}{} & \multicolumn{3}{c|}{COVERAGE} & \multicolumn{3}{c|}{DSO-1}\\
\hline
\rowcolor{lightgray}\multicolumn{1}{|c|}{Anti-Forensics Method Name} & PSNR & SSIM & BRISQUE & PSNR & SSIM & BRISQUE\\
\hline
    Raw Images & - & - & 18.96 & - & - & 13.58\\
    \hline
    Blur\&Sharp & 32.69 & 0.940 & 37.15 & \underline{38.92} & 0.951 & \underline{24.10}\\
    \hline
    Downsize\&Upsize & 32.08 & 0.944 & \underline{34.26} & \underline{39.70} & \underline{0.964} & 30.97\\
    \hline
    Gaussian Noise\&Denoise (Sigma=15) & \underline{37.35} & \underline{0.961} & 36.04 & 37.72 & 0.926 & 35.92\\
    \hline
    Gaussian Noise\&Denoise (Sigma=25) & 34.91 & 0.942 & 39.20 & 36.03 & 0.906 & 43.67\\
    \hline
    JPEG Compression\&JPEG CAR (QF=50) & 35.68 & 0.955 & 38.87 & 37.29 & 0.925 & 41.69\\
    \hline
    JPEG Compression\&JPEG CAR (QF=70) & \underline{37.30} & \underline{0.966} & \underline{35.71} & 38.65 & 0.940 & 33.29\\
    \hline
    Downscale\&Upscale & - & - & - & 38.61 & \underline{0.956} & \underline{30.70}\\
    \hline
\end{tabular}
\caption{Image Quality Metrics For Anti-Forensics Methods (best two metrics are underlined.)}
\label{tab:1} 
\end{table}

\begin{table}[H]
\setlength\arrayrulewidth{1pt} 
\renewcommand{\arraystretch}{1.4}
\centering
\resizebox{\textwidth}{!}{
\centering
\begin{tabular}{|r|c|c|c|c|c|c|c|c|}
\hline
\rowcolor{lightgray}\multicolumn{1}{|c|}{} & \multicolumn{4}{c|}{Trufor} & \multicolumn{4}{c|}{Early Fusion}\\
\hline
\rowcolor{lightgray}\multicolumn{1}{|c|}{} & \multicolumn{2}{c|}{COVERAGE} & \multicolumn{2}{c|}{DSO-1}& \multicolumn{2}{c|}{COVERAGE} & \multicolumn{2}{c|}{DSO-1}\\
\hline
\rowcolor{lightgray}\multicolumn{1}{|c|}{Anti-Forensics Method Name} & Accuracy & Recall & Accuracy & Recall & Accuracy & Recall & Accuracy & Recall \\
\hline
    Raw Images & 0.690 & 0.43 & 0.945 & 0.95 & 0.770 & 0.74 & 0.885 & 0.95\\
    \hline
    Blur\&Sharp & \underline{0.515} & 0.25 & \underline{0.600} & \underline{0.20} & \underline{0.500} & \underline{0.42} & 0.690 & \underline{0.53}\\
    \hline
    Downsize\&Upsize & \underline{0.510} & 0.21 & 0.840 & 0.77 & \underline{0.520} & 0.46 & 0.725 & 0.81\\
    \hline
    Gaussian Noise\&Denoise(Sigma=15) & 0.575 & 0.27 & 0.645 & 0.35 & 0.620 & 0.52 & 0.680 & \underline{0.53}\\
    \hline
    Gaussian Noise\&Denoise(Sigma=25) & \underline{0.515} & \underline{0.17} & 0.605 & \underline{0.26} & 0.590 & \underline{0.45} & 0.630 & \underline{0.35}\\
    \hline
    JPEG Compression\&JPEG CAR(QF=50) & 0.525 & \underline{0.20} & \underline{0.565} & 0.61 & 0.575 & 0.51 & \underline{0.555} & 0.70\\
    \hline
    JPEG Compression\&JPEG CAR(QF=70) & 0.575 & 0.28 & 0.615 & 0.67 & 0.620 & 0.55 & \underline{0.595} & 0.78\\
    \hline
    Downscale\&Upscale & - & - & 0.730 & 0.47 & - & - & 0.690 & \underline{0.53}\\
    \hline
\end{tabular}}
\caption{Manipulation Detection Metrics of SOTA Models (best two metrics are underlined.)}
\label{tab:2} 
\end{table}

\clearpage
\section{Evaluation}
To measure how effective the proposed methods are, we applied the methods on the DSO-1 and COVERAGE datasets and transformed the images. Due to poor results of Downscale\&Upscale method on COVERAGE, its discluded from evaluation. Transformed images are stored with lossless file formats. We computed SSIM, PSNR and BRISQUE quality metrics for the transformed images  and then had the Trufor and Early Fusion models predict the transformed images. Early Fusion scales the images to 2048 pixels on the longest side when predicting, but we did not preprocess the images to make the prediction fair (this is the reason why the Early Fusion model obtained different scores on the DSO-1 dataset than the original paper). When evaluating the methods, we preferred detection over localization because localization results make sense with detection. In addition, it is not effective to examine the localization map of each image individually when a large number of images' authenticity is questioned. 

\begin{figure}[H]
    \centering
    \includegraphics[width=0.65\textwidth]{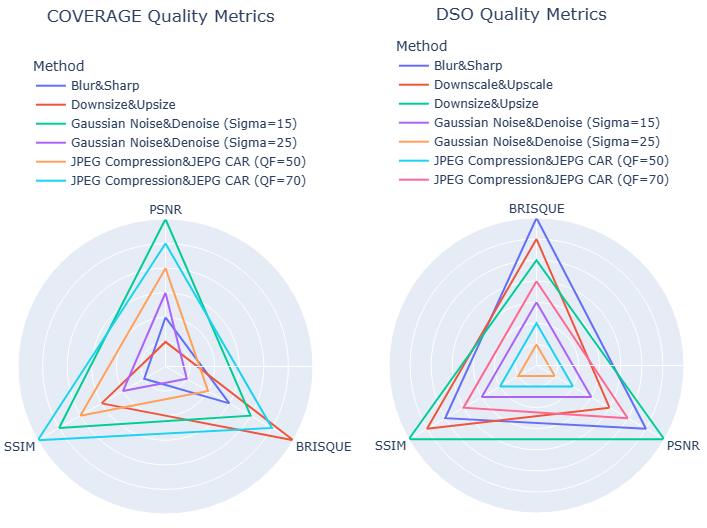}
    \caption{Radar plots of image quality metrics w.r.t. their order(The point further away from the center corresponds to a more successful metric value).}
    \label{fig:2}
\end{figure}

We first analyzed the images qualitatively, and found that when looking at most of the converted images without zooming in, it was difficult to tell the difference between them and the original images. Image quality metrics show decreasing the quality factor and increasing the noise sigma negatively affected the quality (see Table \ref{tab:1}, Figure\ref{fig:2}). It can be said that the image quality decreases when corruption is increased. When the visual comparisons are analyzed (see Figure \ref{fig:5} and \ref{fig:6}), it will be seen that the only method that gives good results in both examples is the JPEG CAR method at 70 QF. Since there are 400 images in total in the DSO-1 and COVERAGE datasets, it is difficult to make general conclusions. From an anti-forensics point of view, it is a serious advantage to have many alternatives that work well in different situations.

\begin{figure}[H]
    \includegraphics[width=1\textwidth]{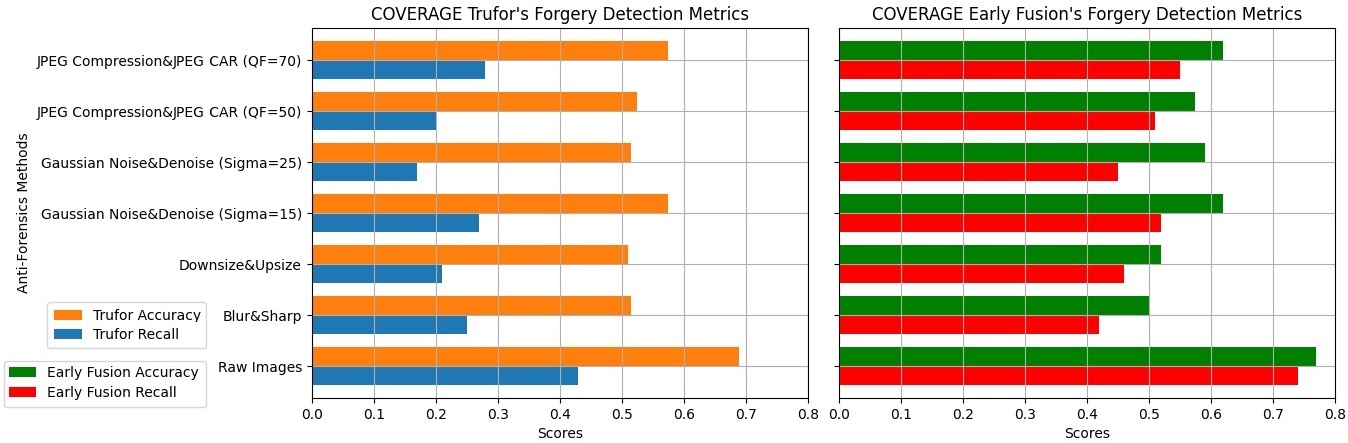}
    \caption{Grouped bar plots of manipulation detection metrics on COVERAGE dataset.}
    \label{fig:3}
\end{figure}

\begin{figure}[H]
    \includegraphics[width=1\textwidth]{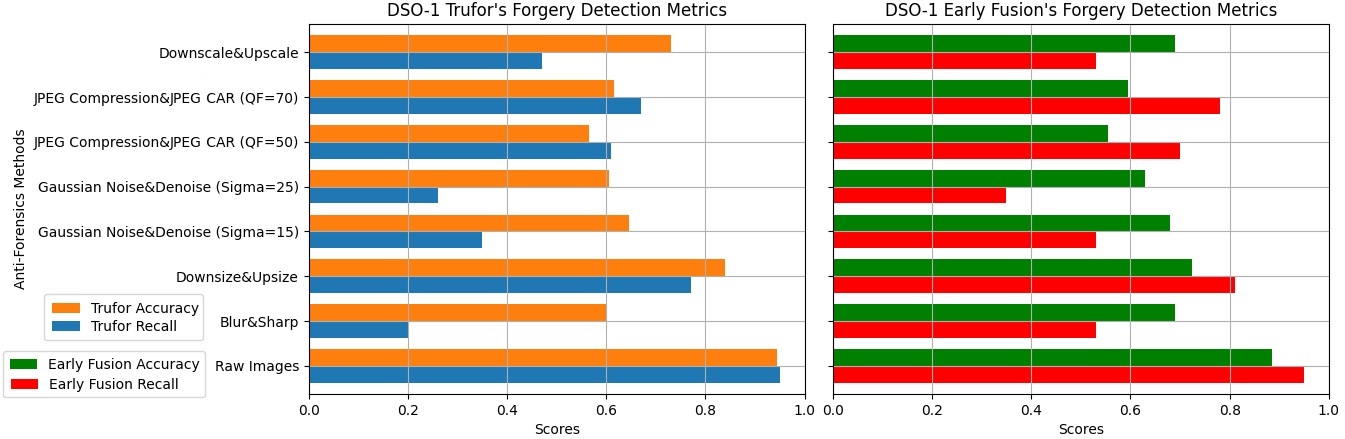}
    \caption{Grouped bar plots of manipulation detection metrics on DSO-1 dataset.}
    \label{fig:4}
\end{figure}

\begin{figure}[H]
    \includegraphics[width=1\textwidth]{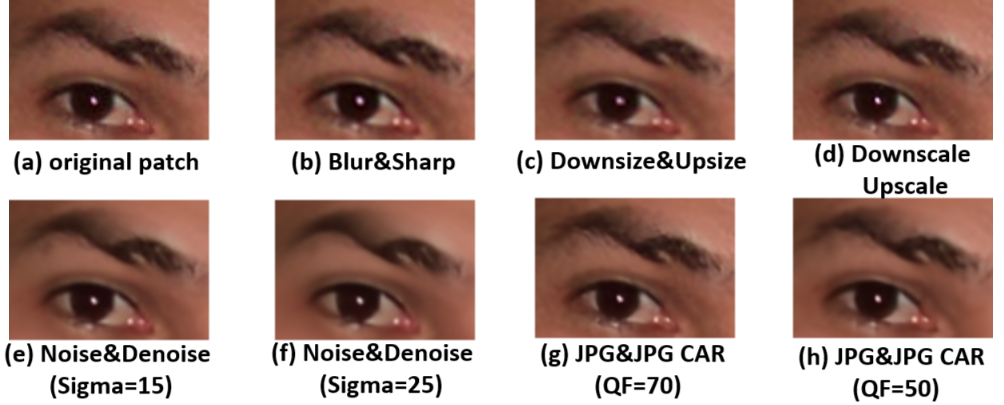}
    \caption{Visual comparison of postprocessing attacks on an image patch(100x70) from DSO-1.}
    \label{fig:5}
\end{figure}

\begin{figure}[H]
    \includegraphics[width=1\textwidth]{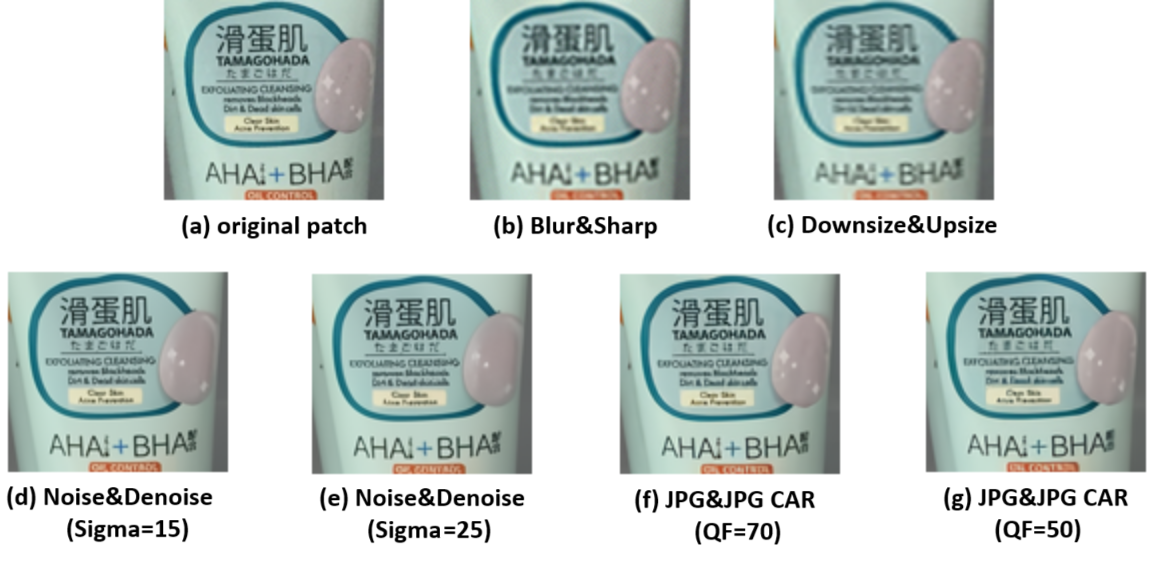}
    \caption{Visual comparison of postprocessing attacks on an image patch(110x100) from COVERAGE.}
    \label{fig:6}
\end{figure}

Table \ref{tab:2}, Figure \ref{fig:3} and \ref{fig:4} shows the effect of anti-forensics methods on image forgery detection. Early Fusion seems to be more resistant than Trufor. This is an expected result since it uses various featurers. The proposed methods reduced the Accuracy of the models towards 50\%, which is the success rate of random prediction. The two best detection models gave poor results against the proposed methods. The reason for the poor results is that none of the existing datasets used advanced anti-forensics methods similar to our proposed ones. Advanced methods were also not used in the data-augmentation phase of the model training.

If the superiority of models with and without deep learning is analyzed, it can be said that both types are successful in reducing score metrics. When choosing the method to be used by the manipulator, priority can be given to image quality, as which method preserves image quality better varies from situation to situation.

\section{Conclusion}
The weakness of SOTA image manipulation detection models to deep image restoration-supported anti-forensics methods has been clearly demonstrated. The use of such anti-forensics methods is necessary to make models resistant to real-world image manipulations. We believe that the inclusion of anti-forensics methods in the data creation and model training processes will take the field of image forgery detection several steps further.

\bibliographystyle{unsrt}  
\bibliography{references}

\end{document}